\documentclass[10pt,twocolumn,letterpaper]{article}

\usepackage[pagenumbers]{iccv} 

\newcommand{\tabtwoobj}{
\begin{table}[t]
    \centering
    \vspace{-.5em}
    \begin{tabular}{l cc ccc}
    \toprule
    Method &&
    & $\mathcal{D}_C$ (↓) & $\mathcal{D}_I$ (↓)
    & $\mathcal{T}_2$ (↑)
    \\ \midrule \midrule
         SDXL~\cite{sdxl} &&
         & 1.09 & 1.21 & 0.80 \\
         MoCE~\cite{moce} &&
         & 0.55 & 0.79 & 0.86 \\
         \sdt~\cite{sdtt} &&
         & 0.66 & 0.75 & 0.84 \\
         \dall~\cite{dalle} &&
         & 0.37 & 0.65 & 0.85 \\
         \sana~\cite{sana} &&
         & 0.46 & 0.74 & 0.86 \\
         \our &&
         & {\bf 0.24} & {\bf 0.62} & {\bf 0.91}
    \\ \bottomrule
    \end{tabular}
    \vspace{-.5em}
    \caption{The comparison on counterfactual T2I with 2 concepts of entities. The best results are in boldface.}
    \vspace{-1.2em}
    \label{tab:2ent}
\end{table}
}

\newcommand{\tabmore}{
\begin{table*}[]
    \centering
    \vspace{-.5em}
    \resizebox{\textwidth}{!}{
    \begin{tabular}{l c ccc c ccc c ccc}
    \toprule
    Method &
    & $\mathcal{V}_3(C)$ ↓ & $\mathcal{V}_3(I)$ ↓
    & $\mathcal{T}_3$ ↑ &
    & $\mathcal{V}_5(C)$ ↓ & $\mathcal{V}_5(I)$ ↓
    & $\mathcal{T}_5$ ↑ &
    & $\mathcal{V}_{mix}(C)$ ↓ & $\mathcal{V}_{mix}(I)$ ↓
    & $\mathcal{T}_{mix}$ ↑
    \\ \midrule \midrule
         SDXL~\cite{sdxl} &
         & 1.69       & 1.83       & 0.64 &
         & 1.78       & 2.02       & 0.12 &
         & 1.48       & 1.69       & 0.48 \\
         MoCE~\cite{moce} &
         & 0.93       & 1.17       & 0.72 &
         & 1.67       & 1.95       & 0.32 &
         & 1.08       & 1.39       & 0.56 \\
         \sdt~\cite{sdtt} &
         & 0.58       & 0.89       & 0.78 &
         & 1.24       & 1.59       & 0.35 &
         & 0.86       & 1.29       & 0.55 \\
         \dall~\cite{dalle} &
         & 0.67       & 0.76       & 0.76 &
         & 1.58       & 1.10       & 0.36 &
         & 0.75       & 1.48       & 0.60 \\
         \sana~\cite{sana} &
         & 0.63       & 0.80       & 0.82 &
         & 1.09       & 0.96       & 0.41 &
         & 0.87       & 1.03       & 0.61 \\
         \our &
         & {\bf 0.54} & {\bf 0.68} & {\bf 0.84} &
         & {\bf 0.95} & {\bf 0.92} & {\bf 0.45} &
         & {\bf 0.74} & {\bf 0.96} & {\bf 0.62} \\
         \bottomrule
    \end{tabular}
    }
    \vspace{-.5em}
    \caption{The score evaluation on counterfactual T2I with 3, 5 and mixed numbers of concepts of entities. The best results are in boldface.}
    \vspace{-1em}
    \label{tab:more}
\end{table*}
}

\newcommand{\tabthrii}{
\begin{table}[h]
    \centering
    \begin{tabular}{c cc ccc}
    \toprule
    Threshold &&
    & \dc (↓) & \di (↓)
    & $\mathcal{T}_{2}$ (↑)
    \\ \midrule \midrule
    20\% &&
    & 1.98       & 1.92       & 0.32 \\
    40\% &&
    & 1.89       & 1.79       & 0.36 \\
    60\% &&
    & {\bf 0.24} & {\bf 0.62} & {\bf 0.91} \\
    80\% &&
    & 1.82       & 1.68       & 0.46 \\
    100\% &&
    & -       & -       & - \\
    dynamic &&
    & 0.86 & 0.72 & 0.77
    \\ \bottomrule
    \end{tabular}
    \vspace{-.5em}
    \caption{The performance on 2-counterfactual problems under different threshold values in question blocks' pass check. The best results are in boldface. We can see fixed 60\% is the best choice and dynamic method outperforms many fixed ways. The 100\% method turns to collapse in loop with no results.}
    \vspace{-0.8em}
    \label{tab:threshold2}
\end{table}
}

\newcommand{\tabthrmore}{
\begin{table*}[t]
    \centering
    \vspace{-40em}
    \resizebox{\textwidth}{!}{
    \begin{tabular}{c cc ccc c ccc c ccc}
    \toprule
    Threshold &&
    & $\mathcal{V}_{3}(C)$ ↓ & $\mathcal{V}_{3}(I)$ ↓
    & $\mathcal{T}_{3}$ ↑ &
    & $\mathcal{V}_{5}(C)$ ↓ & $\mathcal{V}_{5}(I)$ ↓
    & $\mathcal{T}_{5}$ ↑ &
    & $\mathcal{V}_{mix}(C)$ ↓ & $\mathcal{V}_{mix}(I)$ ↓
    & $\mathcal{T}_{mix}$ ↑
    \\ \midrule \midrule
    20\% &&
    & 1.87       & 1.74       & 0.29 &
    & 2.01       & 1.78       & 0.26 &
    & 1.91       & 1.83       & 0.27 \\
    40\% &&
    & 1.56       & 1.48       & 0.49 &
    & 1.56       & 1.42       & 0.30 &
    & 1.66       & 1.53       & 0.39 \\
    60\% &&
    & {\bf 0.54} & {\bf 0.68} & {\bf 0.84} &
    & {\bf 0.95} & {\bf 0.92} & {\bf 0.45} &
    & {\bf 0.74} & {\bf 0.96} & {\bf 0.62} \\
    80\% &&
    & 0.75       & 0.88       & 0.72 &
    & 1.36       & 1.39       & 0.42 &
    & 1.05       & 1.16       & 0.43 \\
    100\% &&
    & -       & -       & - &
    & -       & -       & - &
    & -       & -       & - \\
    dynamic &&
    & 0.92 & 0.86 & 0.68 &
    & 1.46 & 1.33 & 0.28 &
    & 1.44 & 1.32 & 0.51
    \\ \bottomrule
    \end{tabular}
    }
    \vspace{-.5em}
    \caption{The performance on 3, 5 and mixed-counterfactual problems under different threshold values in question blocks' pass judge. 100\% setting leads to collapses in loop in all situations. The best results are in boldface. We can see fixed 60\% is the best choice while dynamic method outperforms many fixed ways.}
    \label{tab:thresholdmore}
\end{table*}
}

\newcommand{\figoverview}{
\begin{figure*}
    \centering
    \includegraphics[width=\linewidth]{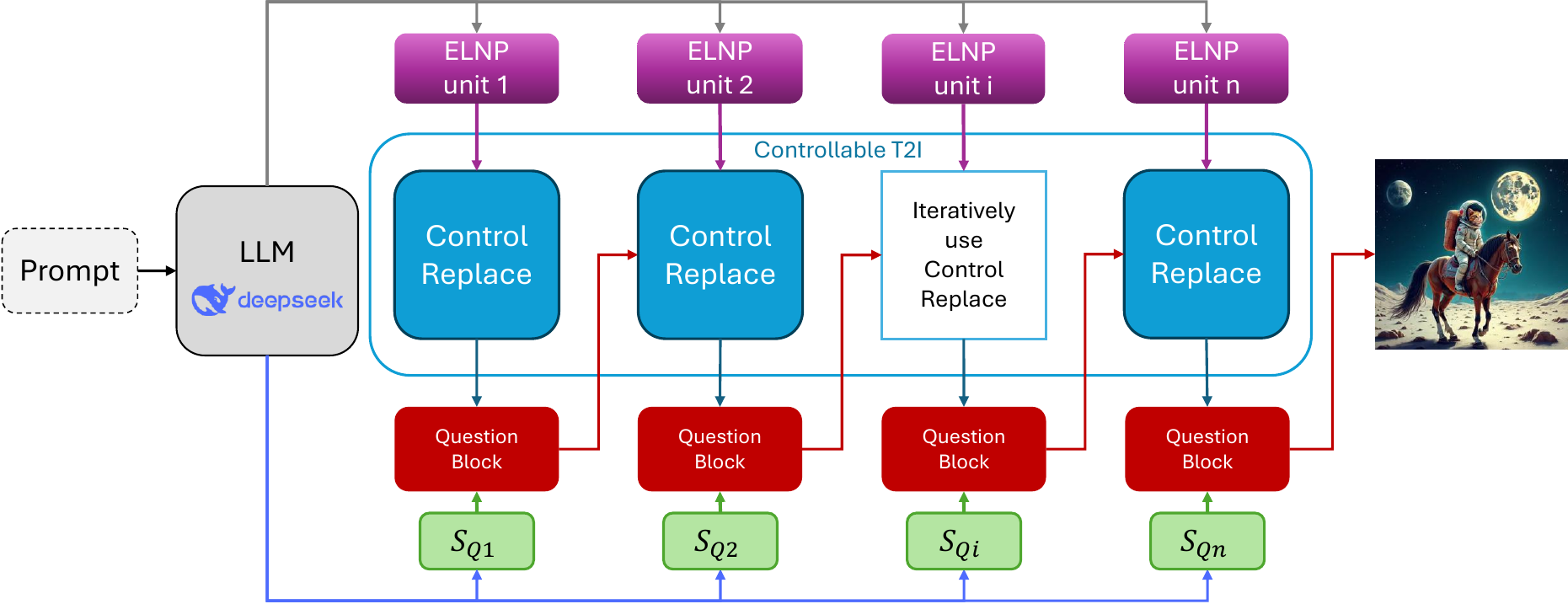}
    \caption{The overview of the ELNP and Question Blocks strategy for counterfactual T2I.}
    \label{fig:overview}
\end{figure*}
}

\newcommand{\figelnpa}{
\begin{figure}
    \centering
    \includegraphics[width=\linewidth]{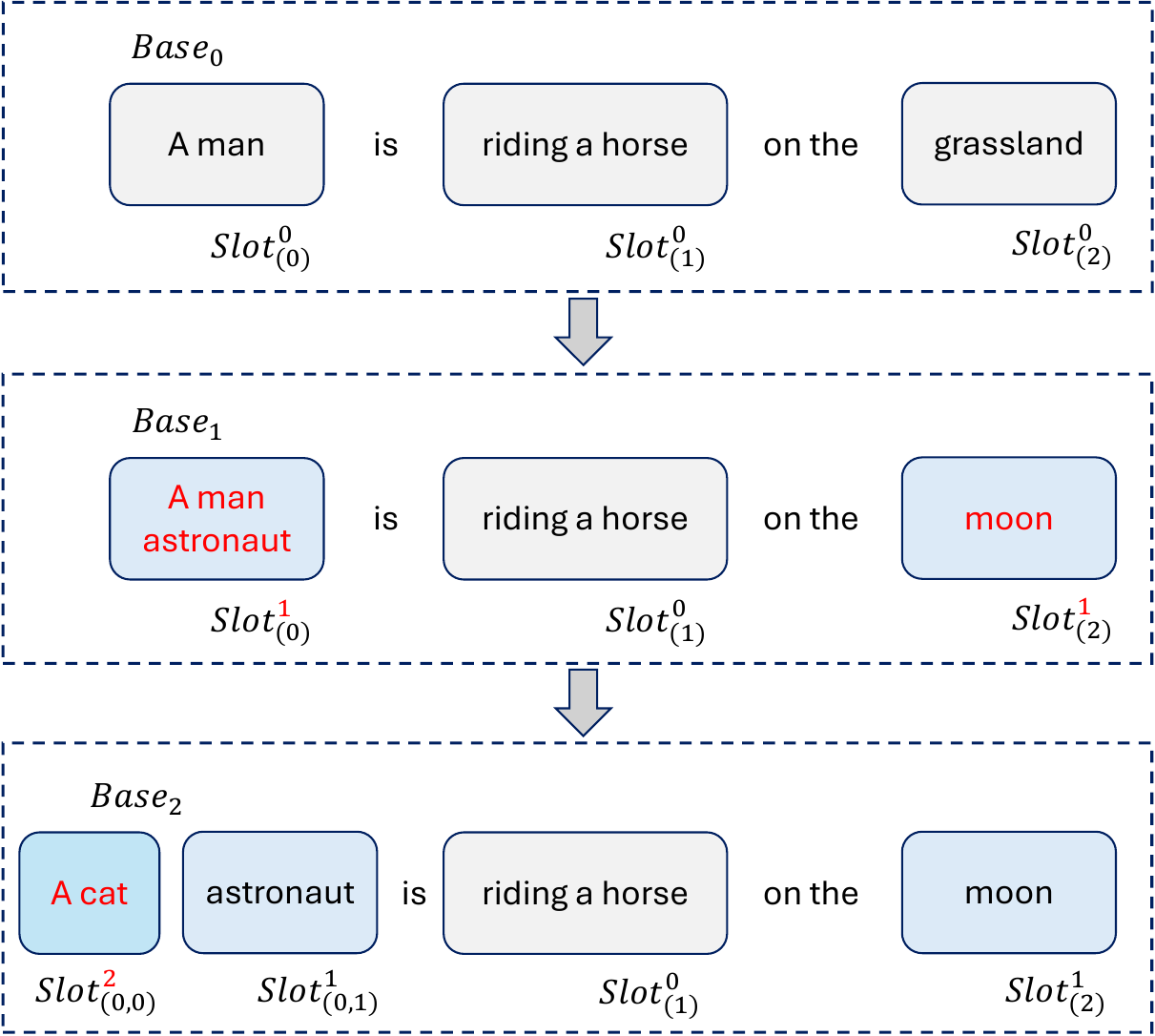}
    \caption{An example of replacement from original base to the target counterfactual text prompt: {\em A cat astronaut is riding a horse on the moon}.}
    \label{fig:elnpa}
\end{figure}
}

\newcommand{\figques}{
\begin{figure}
    \centering
    \includegraphics[width=.9\linewidth]{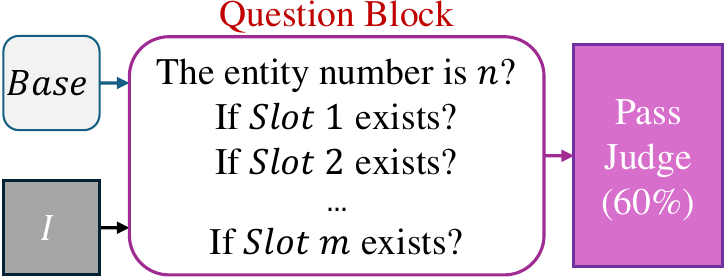}
    \caption{A question block. $I$ is the intermediate image decoded from the latent space. Pass judge will check whether the positive answers cover no less than 60\% of all questions in this block.}
    \label{fig:ques}
\end{figure}
}

\newcommand{\figqual}{
\begin{figure}
    \centering
    \includegraphics[width=\linewidth]{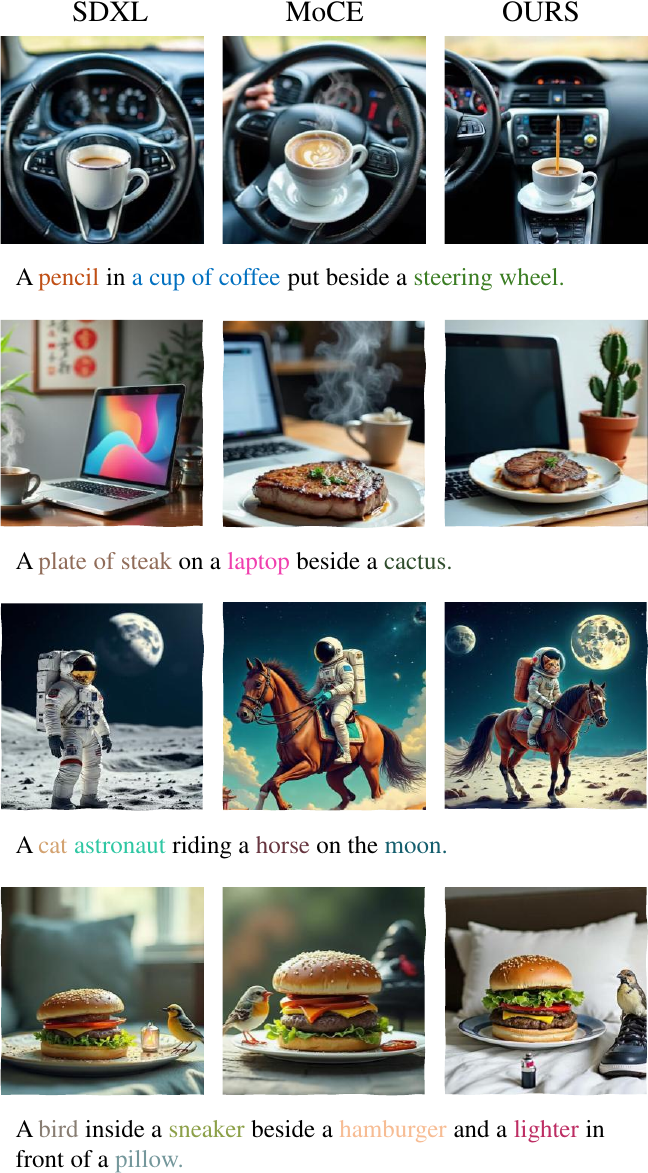}
    \caption{The qualitative examples of some counterfactual text prompts with more than two entities. Both SDXL~\cite{sdxl} and MoCE~\cite{moce} cannot cover every concepts while our method succeeds to fully cover all the concepts mentioned in the prompts.}
    \label{fig:qual}
\end{figure}
}

\newcommand{\promptELNPA}{
\begin{figure}[h]
    \centering
    \vspace{-0.4em}
    \begin{tcolorbox}[fonttitle = \small\bfseries, title=\deepseek-R1,colframe=gray!2!black,colback=gray!2!white,boxrule=1pt,boxsep=0pt,left=5pt,right=5pt,fontupper=\footnotesize, halign title = flush center]
You are a sentence parser, please extract all the objects in the prompt according to the exemplar below:\\
'''\\
Exemplar:\\
Prompt:
\{A cat astronaut is riding a horse running on the moon.\}\\
All the objects:
\{cat, astronaut, horse, moon\}\\
'''\\
Please give all the objects in the Prompt: \{A butterfly is flying over a cup of coke.\}
    \end{tcolorbox}
    \vspace{-1.3em}
    \caption{The prompt template to generate ELNP: extract key entities. In this example, we ask \deepseek-R1 to process {\em a butterfly is flying over a cup of coke}.}
    \label{fig:epre}
\end{figure}
}

\newcommand{\promptELNPB}{
\begin{figure}[h]
    \centering
    \vspace{-1.5em}
    \begin{tcolorbox}[fonttitle = \small\bfseries, title=\deepseek-R1,colframe=gray!2!black,colback=gray!2!white,boxrule=1pt,boxsep=0pt,left=5pt,right=5pt,fontupper=\footnotesize, halign title = flush center]
You are a sentence parser and status analyzer, please extract all the objects in the prompt and the status of each object according to the exemplar below:\\
'''\\
Exemplar:\\
Prompt:
\{A cat astronaut is riding a horse running on the moon.\}\\
All the objects:
\{cat, astronaut, horse, moon\}\\
\\
The status:\\
\{
cat: [astronaut, riding a horse],
astronaut: [cat, riding a horse],\\
horse: [being ridden, running],
moon: [there is a horse running on it]
\}\\
'''\\
Please give all the objects and the status in the Prompt: \{A butterfly is flying over a cup of coke.\}
    \end{tcolorbox}
    \vspace{-1.3em}
    \caption{The prompt template to generate ELNP: analyze the status of key entities.}
    \vspace{-1em}
    \label{fig:eprs}
\end{figure}
}

\newcommand{\promptELNPC}{
\begin{figure}[t]
    \centering
    \begin{tcolorbox}[fonttitle = \small\bfseries, title=\deepseek-R1,colframe=gray!2!black,colback=gray!2!white,boxrule=1pt,boxsep=0pt,left=5pt,right=5pt,fontupper=\footnotesize, halign title = flush center]
You are a prompt splitter. For a prompt that describes a bizarre scene, you should first figure out what is the most ordinary condition of the scene and provide an ordered series of replacing the objects in the ordinary condition to restore the original bizarre description. Give out the ordered series according to the two exemplars below.
\\
'''
\\
Exemplar 1:\\
The bizarre scene:\\
\{A cat is riding a horse running on the grassland.\}\\
\\
Its most ordinary condition:\\
\{A man is riding a horse running on the grassland.\}\\
\\
The series to restore to the original description:\\
1: [(man, cat)]$\sim$[\{A man is riding a horse running on the grassland.\}, \{A cat is riding a horse running on the grassland.\}]\\
\\ 
'''\\
Exemplar 2:\\
The bizarre scene:\\
\{A cat astronaut is riding a horse running on the moon.\}\\
\\
Its most ordinary condition:\\
\{A man is riding a horse running on the grassland.\}\\
\\
The series to restore to the original description:\\
1: [(man, human astronaut), (grassland, moon)]$\sim$[\{A man is riding a horse running on the grassland.\}, \{A human astronaut is riding a horse running on the moon.\}]\\
2: [(human, cat)]$\sim$[\{A human astronaut is riding a horse running on the moon.\}, \{A cat astronaut is riding a horse running on the moon.\}]\\
'''\\
\\
Please give the ordered series of the bizarre scene:
\{A butterfly is flying over a cup of coke.\}
    \end{tcolorbox}
    \vspace{-1.4em}
    \caption{The prompt template to generate ELNP: order of replacements. Based on the previous parsing toward entities in the exemplar, we can give clear instruction exemplars for \deepseek to generate the ordered series instructing the controllable model to replace entities within the image step-by-step.}
    \label{fig:epro}
\end{figure}
}

\newcommand{\figquestionblock}{
\begin{figure}[t]
    \centering
    \vspace{1.6em}
    \begin{tcolorbox}[fonttitle = \small\bfseries, title=\textsc{Question Blocks},colframe=gray!2!black,colback=gray!2!white,boxrule=1pt,boxsep=0pt,left=5pt,right=5pt,fontupper=\footnotesize, halign title = flush center]
The entity number is \{{\color{red} 3}\}?\\
If a \{{\color{orange} horse}\} exists in the image?\\
\tcbline
The entity number is \{{\color{red} 3}\}?\\
If a \{{\color{orange} horse}\} exists in the image?\\
If a \{{\color{teal} astronaut}\} exists in the image?\\
\tcbline
The entity number is \{{\color{red} 3}\}?\\
If a \{{\color{orange} horse}\} exists in the image?\\
If a \{{\color{teal} astronaut}\} exists in the image?\\
If a \{{\color{blue} cat}\} exists in the image?\\
\tcbline
The entity number is \{{\color{red} 3}\}?\\
If a \{{\color{orange} horse}\} exists in the image?\\
If a \{{\color{teal} astronaut}\} exists in the image?\\
If a \{{\color{blue} cat}\} exists in the image?\\
If a \{{\color{violet} moon}\} exists in the image?
    \end{tcolorbox}
    \vspace{-1.3em}
    \caption{The questions for each period can be generated from the key entities and the order.}
    \vspace{-0.6em}
    \label{fig:qb}
\end{figure}
}

\usepackage{CJKutf8}
\usepackage[most]{tcolorbox}

\usepackage[T1]{fontenc}

\usepackage{amssymb}
\usepackage{pifont}

\usepackage[utf8]{inputenc}

\usepackage{microtype}

\usepackage{inconsolata}

\usepackage{graphicx}
\usepackage{xspace}

\tcbset{
    userstyle/.style={
        enhanced,
        colback=white,
        colframe=black,
        colbacktitle=gray!20,
        coltitle=black,
        rounded corners,
        sharp corners=north,
        boxrule=0.5pt,
        drop shadow=black!50!white,
        attach boxed title to top left={
            xshift=-2mm,
            yshift=-2mm
        },
        boxed title style={
            rounded corners,
            size=small,
            colback=gray!20
        }
    },
    replystyleg/.style={
        enhanced,
        colback=green!15,
        colframe=black,
        colbacktitle=green!30,
        coltitle=black,
        boxrule=0.5pt,
        drop shadow=black!50!white,
        rounded corners,
        sharp corners=north,
        attach boxed title to top right={
            xshift=-2mm,
            yshift=-2mm
        },
        boxed title style={
            rounded corners,
            size=small,
            colback=green!40
        }
    },
    replystyler/.style={
        enhanced,
        colback=red!15,
        colframe=black,
        colbacktitle=red!40,
        coltitle=black,
        boxrule=0.5pt,
        drop shadow=black!50!white,
        rounded corners,
        sharp corners=north,
        attach boxed title to top right={
            xshift=-2mm,
            yshift=-2mm
        },
        boxed title style={
            rounded corners,
            size=small,
            colback=red!40
        }
    }
}

\newcommand{\gpt}{\mbox{\textsc{GPT-3.5}}\xspace}

\newcommand{\deepseek}{\mbox{\textsc{DeepSeek}}\xspace}

\newcommand{\sdt}{\mbox{SD3-medium}\xspace}
\newcommand{\dall}{\mbox{DALL$\cdot$E3}\xspace}
\newcommand{\sana}{\mbox{\textsc{Sana-1.6B}}\xspace}
\newcommand{\our}{\mbox{RIT (ours)}\xspace}

\newcommand{\D}[1]{\mbox{$\mathcal{D}_{#1}$}\xspace}
\newcommand{\dc}{\mbox{\D{C}}\xspace}
\newcommand{\di}{\mbox{\D{I}}\xspace}

\newtcolorbox{querygpt}[1][]{
    userstyle,
    title=\gpt
}

\definecolor{iccvblue}{rgb}{0.21,0.49,0.74}
\usepackage[pagebackref,breaklinks,colorlinks,allcolors=iccvblue]{hyperref}

\def\paperID{13732} 
\def\confName{ICCV}
\def\confYear{2025}

\title{Replace in Translation: Boost Concept Alignment in Counterfactual Text-to-Image}

\author{Sifan Li\textsuperscript{1} \quad
Ming Tao\textsuperscript{2} \quad 
Hao Zhao\textsuperscript{3} \quad
Ling Shao\textsuperscript{4} \quad
Hao Tang\textsuperscript{5} \\
\textsuperscript{1}Liaoning University \quad
\textsuperscript{2}Nanjing University of Posts and Telecommunications \quad  
\textsuperscript{3}Tsinghua University \\
\textsuperscript{4}University of Chinese Academy of Sciences \quad
\textsuperscript{5}Peking University \\
{\tt\small \href{mailto:haotang@pku.edu.cn}{haotang@pku.edu.cn}} \\
}

\begin{document}
\maketitle
\begin{abstract}
Text-to-Image (T2I) has been prevalent in recent years, with most common condition tasks having been optimized nicely. Besides, counterfactual Text-to-Image is obstructing us from a more versatile AIGC experience. For those scenes that are impossible to happen in real world and anti-physics, we should spare no efforts in increasing the factual feel, which means synthesizing images that people think very likely to be happening, and concept alignment, which means all the required objects should be in the same frame. In this paper, we focus on concept alignment. As controllable T2I models have achieved satisfactory performance for real applications, we utilize this technology to replace the objects in a synthesized image in latent space step-by-step to change the image from a common scene to a counterfactual scene to meet the prompt. We propose a strategy to instruct this replacing process, which is called as Explicit Logical Narrative Prompt (ELNP), by using the newly SoTA language model DeepSeek to generate the instructions. Furthermore, to evaluate models' performance in counterfactual T2I, we design a metric to calculate how many required concepts in the prompt can be covered averagely in the synthesized images. The extensive experiments and qualitative comparisons demonstrate that our strategy can boost the concept alignment in counterfactual T2I.
\end{abstract}

\section{Introduction}
\label{sec:intro}
\figqual
Recent advancements in Text-to-Image (T2I) generation have revolutionized content creation~\cite{cogview,mansimov,photo,consistency}, enabling high-fidelity synthesis of images from textual prompts~\cite{mirrorgan,vqvae,glide,ldm,sdkind,sdxl,dalle,sana}. While state-of-the-art models excel at generating realistic scenes for common scenarios~\cite{ddpm,ddim,ddpm2,guideddiff,dit,minidalle3}, their performance degrades significantly when tasked with counterfactual generation—synthesizing images of improbable or physically implausible scenarios (\eg, "a cat astronaut riding a horse on the moon"). Such counterfactual T2I demands not only visual plausibility but also strict adherence to concept alignment, where all specified entities must coexist coherently in the generated image~\cite{moce}. Despite progress in controllable T2I frameworks~\cite{controlgan,addingcond,sun2024anycontrolcreateartworkversatile,insp2p}, existing methods struggle to balance multiple conflicting concepts, often omitting entities or blending them unrealistically~\cite{rich1,rich2,rich3}. This limitation hinders applications in creative design, educational visualization, and synthetic data generation, where complex scenes are essential~\cite{grain,refact}.

Current approaches to counterfactual T2I, such as causal disentanglement or fine-tuning with rare tokens, primarily address single-concept edits (\eg, swapping object attributes) but fail to scale to multi-entity scenarios~\cite{benchdalle3,pair,drtune}. For instance, generating a scene with five unrelated entities frequently results in partial omissions or semantic collisions~\cite{moce,infthe}. Furthermore, evaluation metrics for counterfactual tasks remain underdeveloped~\cite{testing}, with existing tools like Multi-Concept Disparity~\cite{moce} limited to pairwise interactions. These gaps underscore the need for a systematic strategy to guide multi-concept integration and a robust evaluation framework to quantify alignment.

In this work, we propose
\textsc{{\bf R}eplace {\bf I}n {\bf T}ranslation} ({\bf RIT}),
a novel paradigm for enhancing concept alignment in counterfactual T2I. Our approach leverages the observation that counterfactual scenes can often be decomposed into iterative modifications of a plausible base scene. For example, transforming "a man riding a horse on grassland" into "a cat astronaut riding a horse on the moon" involves stepwise replacements: first altering the background ("grassland → moon"), then the subject ("man → cat astronaut"). To operationalize this intuition, we introduce the Explicit Logical Narrative Prompt (ELNP), a structured workflow that directs controllable T2I models (\eg, ControlNet~\cite{addingcond}) to replace entities in latent space iteratively. We utilize the new powerful language model \deepseek-R1~\cite{deepseek} to generate the ELNP instructions of steps to replace. At each step, a question block validates the presence of required concepts, preventing error propagation by reverting to prior states if checks fail. Additionally, we design two novel metrics—Multi-Concept Variance and Targeted Entities Coverage—to evaluate alignment and completeness in multi-entity counterfactual scenes.

Our contributions can be summarized as follows:
\begin{itemize}
    \item {
    {\bf Method}: We propose ELNP, a strategy for hierarchical entity replacement in latent space, coupled with iterative validation through question blocks, to enhance concept alignment in complex counterfactual scenes.
    }
    \item {
    {\bf Benchmark}: We extend the LC-Mis dataset~\cite{moce} to construct a comprehensive benchmark for multi-entity (2–5 concepts) and mixed-counterfactual scenarios, enabling systematic evaluation.
    }
    \item {
    {\bf Metrics}: We devise Multi-Concept Variance and Targeted Entities Coverage to quantify alignment and completeness, addressing limitations in existing pairwise metrics.
    }
    \item {
    {\bf Empirical Validation}: Extensive experiments demonstrate state-of-the-art performance across 2–5 entity tasks, with 91\% entity coverage for 2-concept scenes and 45\% for highly complex 5-concept scenarios, significantly outperforming baselines like SDXL~\cite{sdxl}, \sdt~\cite{sdtt}, \dall~\cite{dalle} and \sana~\cite{sana}.
    }
\end{itemize}

By bridging the gap between controllable synthesis and counterfactual reasoning, our work advances the frontier of T2I generation, enabling richer, more reliable creation of imaginative visual content. Code, datasets, and visual results will be made publicly available to support further research.


\section{Related Works}
\label{sec:rel}

This section reviews advancements in counterfactual text-to-image (T2I) generation and controllable T2I synthesis, focusing on their methodologies, challenges, and contributions to the field.

\subsection{Counterfactual Text-to-Image}

Counterfactual reasoning in T2I models addresses "what-if" scenarios by generating images that deviate from factual conditions while preserving logical consistency. A pivotal approach involves Generative Causal Models (GCMs), which disentangle class-agnostic features (\eg, pose) and class-specific attributes (\eg, object traits) to synthesize counterfactual samples. For instance, Zhang~\etal~\cite{yue2021} introduced a framework where replacing class attributes (\eg, "bird species") while retaining sample-specific features (\eg, posture) generates plausible counterfactual images. Their method employs contrastive and adversarial losses to enforce faithfulness between generated and original distributions, enabling applications in zero-shot and open-set recognition.

Recent work further integrates counterfactual faithfulness into diffusion models. For example, DreamBooth~\cite{dreambooth} fine-tunes pre-trained T2I models using rare token identifiers to bind specific subjects, ensuring that counterfactual edits (\eg, altering backgrounds or styles) preserve core identity features. This approach mitigates overfitting by combining self-generated regularization data with semantic alignment losses, enhancing robustness in multi-context generation~\cite{he2024dataperspectiveenhancedidentity}.

\subsection{Controllable Text-to-Image}

Controllable text-to-image (T2I) generation~\cite{2019controllable} focuses on integrating diverse spatial, semantic, or stylistic conditions to guide image synthesis.
\vspace{.2em}
\\
{\bf Multi-Control Frameworks}. Methods like AnyControl~\cite{sun2024anycontrolcreateartworkversatile} unify spatial signals (\eg, depth maps, sketches) and textual prompts through a Multi-Control Encoder. By alternating fusion and alignment blocks, it harmonizes complex relationships between conditions, addressing challenges like input flexibility and semantic-textual compatibility. Similarly, ControlNet~\cite{addingcond} extends T2I models by encoding auxiliary conditions (\eg, edges, segmentation maps) into latent representations, enabling fine-grained structural control.
\vspace{.2em}
\\
{\bf Prompt Optimization and Style Adaptation}. NeuroPrompts~\cite{neuroprompt} automates prompt engineering using reinforcement learning, aligning user inputs with diffusion-model-friendly language. For style-aware generation, Magic Insert~\cite{magicinsert} combines LoRA-based style infusion with domain-adapted object insertion, enabling realistic drag-and-drop of subjects into stylized scenes while preserving identity.
\vspace{.2em}
\\
{\bf Concept Editing and Personalization}.
EMCID~\cite{emcid24} tackles large-scale concept editing (\eg, updating 1,000+ concepts) via dual self-distillation and closed-form weight adjustments, ensuring minimal quality degradation. DreamBooth~\cite{dreambooth} and Textual Inversion~\cite{magicinsert} personalize models by binding rare tokens to specific subjects, though challenges persist in preserving fine details like logos or text~\cite{he2024dataperspectiveenhancedidentity}.
\vspace{.2em}
\\
{\bf Sketch and Structure Guidance}.
Training-free methods like Latent Optimization~\cite{ding2024trainingfreesketchguideddiffusionlatent} leverage cross-attention maps to align generated images with input sketches, bridging the gap between abstract drawings and photorealistic outputs.
\vspace{.2em}

While counterfactual methods excel in hypothetical generation, they often rely on disentanglement assumptions that may not hold for complex real-world data~\cite{yue2021,he2024dataperspectiveenhancedidentity}. Controllable T2I models, though versatile, face trade-offs between flexibility (\eg, supporting arbitrary condition combinations~\cite{sun2024anycontrolcreateartworkversatile}) and computational efficiency. Future work may explore hybrid frameworks that unify causal reasoning with multi-modal control, enhancing both generalization and user-directed creativity.

\section{Benchmark}
\label{sec:benchmark}

In this section, we detail the process of constructing the datasets and designing the new evaluation metric for multiple entities counterfactual T2I, which addresses limitations in prior frameworks.




\subsection{Data Construction}

We develop datasets for counterfactual scenes involving more than two entities by extending the LC-Mis dataset~\cite{moce}, which originally targets textual prompts containing two concepts of entities that are statistically unlikely to co-occur in a single scene. Formally, we define an $n$-counterfactual problem as a text-to-image generation task where the input text specifies $n$ distinct entities with minimal real-world co-occurrence likelihood. The LC-Mis dataset, by this definition, addresses 2-counterfactual problems.

To advance research in complex counterfactual reasoning, this work focuses on evaluating scenarios with $n>2$ entities. Consequently, we augment the original dataset by generating concept combinations with higher cardinality (\eg, 3, 5, and mixed-entity configurations). This extension enables systematic benchmarking of multi-entity alignment in counterfactual T2I generation, addressing limitations in prior datasets constrained to pairwise entity interactions.

\subsubsection{Concept Distance}

We first extract all the concepts contained in the old dataset and define them as a list $Concepts = \{concept_0, concept_1,...,concept_i,...,concept_n\}$.
With the advances of large language models today, we give each two of the concepts ($concept_i$ and $concept_j$ where $i \ne j$) and ask the language model to calculate the probability $p$ of the two given entities appearing in the same scene. The quantized probability $(1-p)$ is named as the distance between two concepts, whereas $p$ is lower, the two entities are more unlikely to appear in the same scene denoting a larger distance. If we have $n$ concepts, we can obtain $n(n-1)$ items of distance between any two of the concepts.
When we want to construct a concept collection with $m$ entities to tackle $m$-counterfactual problems, we can calculate the total distance of these $m$ concepts. For each pair in these $m$ concepts, we calculate the distance between them and sum every distance between each pair to represent the total distance. The number of pairs in $m$ concepts should be $K=\tbinom{m}{2}=\frac{m!}{2!(m-2)!}$ and the total distance should be $D^m=\sum_{i \ne j}^{K}{distance(concept_i, concept_j)}$. Among all these $n$ concepts, we can extract all the collections with $m$ different concepts and name them collection-m. The total number of all collections-m should be $M=\tbinom{n}{m}=\frac{n!}{m!(n-m)!}$ in total. Thereafter, every collections-m takes its own total distance $D^m_i$.
For example, if we want to construct a collection-3, we extract three different concepts from $Concepts$, sum all three distances between each two of them as the total distance $D^3$.

\subsubsection{Counterfactual Scene Text Generation}

For one $m$-counterfactual problem, we should construct a dataset containing multiple prompts that describe a scene with $m$ concepts of entities. For a $mixed$-counterfactual problem, we should construct a dataset containing prompts that describe scenes with different numbers of concepts of entities. As aforementioned, for $m$-counterfactual problem, we first extract all collections-m, calculate $D^m$ for each of them, and then rank them by $D^m$ in descending order where a higher rank means a more counterfactual scene. We pick the first $r$ of them, if we want to generate $r$ prompts that contain $m$ concepts of entities. For all these picked collections that can be used to generate the most counterfactual scenes, we tell language models to connect the concepts in the collection with simple words or just commas and 'and'. For example, if we collect ['cat', 'hat', 'book'], the language model can generate {\em cat, hat and book} or {\em cat in hat sits beside a book}. We choose the latter pattern to generate prompts. Compared to simple comma connection, the natural sentence descriptions are more close to real text-to-image. Mostly, we do not simply pile words to synthesize images.

Based on the 126 concept pairs in LC-Mis dataset. We construct subsets for $2,3,5$ and mixed-counterfactual problems, named as dataset-2, 3, 5 and mixed. Each of them contains 120 concepts collections. For dataset-mixed, we pick 40 highest ranked collections in each of datasets-2, 3 and 5. We utilize \deepseek as the language model to generate the counterfactual scene prompts for each collection in these datasets.

\subsection{Quantize Counterfactual Scenes}
\figoverview
Recent studies have proposed several evaluation metrics for counterfactual text-to-image (T2I) generation, such as Multi-Concept Disparity~\cite{moce}. While this metric effectively evaluates scenes involving two entities, its applicability remains limited to pairwise counterfactual scenarios, necessitating extensions to address more complex configurations with three or more entities.
Furthermore, inherent limitations in the metric—such as its inability to robustly handle multi-entity alignment and its susceptibility to misinterpretations of concept interactions—result in suboptimal performance when assessing T2I outputs for high-order counterfactual tasks. These shortcomings underscore the need to revise existing frameworks and develop a more comprehensive evaluation metric tailored to multi-concept scenarios.

\subsubsection{Multi-Concept Variance}

Multi-Concept Disparity is a metric designed to evaluate the alignment between text and images containing multiple concepts in a scene. When a specific score, such as $ClipScore$~\cite{clipscore} denoted as $S_c$, is used to assess a particular feature of text-to-image (T2I) generation—for example, the fidelity of the generated image $I$ to a given concept $\mathcal{A}$—the calculation can be expressed as:
\begin{equation}
    S_c(I,\mathcal{A})=ClipScore(I,\mathcal{A}).
\end{equation}
Sole reliance on ClipScore is insufficient to accurately capture the correlation between the final image $I$ and the entity $\mathcal{A}$. To address this limitation, the metric was refined to incorporate the ClipScore of the final image for both concepts and their associated textual descriptions generated by \gpt. This revised approach can be formalized as follows: 
\begin{equation}
    S(I, \mathcal{A})=Max(S_c(I,\mathcal{A}), S_c(I,\mathcal{A}_{description})).
    \label{eq:S}
\end{equation}
This modification enhances the entity's ClipScore while mitigating misalignment effects to a certain extent. Using this score $S$, the Multi-Concept Disparity denoted as $\mathcal{D}$ can be formally calculated as follows:
\begin{equation}
    \mathcal{D}=|S(I,\mathcal{A})-S(I,\mathcal{B})|.
    \label{eq:D}
\end{equation}
For 2-counterfactual problems, $\mathcal{D}$ presents the difference in the scores of an image between two concepts. A larger $\mathcal{D}$ denotes a higher likelihood that one concept within it becomes blurred.

Naive subtraction cannot be applied for more than two concepts.
To expand Multi-Concept Disparity, we need to update the method how it calculates the difference. In reality, $\mathcal{D}$ can be regarded as the score of balancing the two concepts. When $\mathcal{D}$ turns lower, $S(I,\mathcal{A})$ is more similar to $S(I,\mathcal{B})$. Therefore, we can use biased variance to simulate this disparity score in more than two concepts, which is named Multi-Concept Variance. For $n$ concepts, when we calculate a specific score like $ClipScore$, we use the process in~\cref{eq:S}, and calculate Multi-Concept Variance $\mathcal{V}_n$ as:
\begin{equation}
    \mathcal{V}_n=\frac{1}{\sqrt{n}}{\sum_{i}^{n}{(S(I,concept_i)-\bar{S})^2}},
    \label{eq:V}
\end{equation}
where $\bar{S}$ denotes the mean value of all $S(I,concept_i)$.
We denote the score with $\sim ClipScore$ to indicate what score is specifically used as $S$.
For mixed-counterfactual problems, if there are $k$ different kinds of counterfactual problems are mixed, and their total numbers are $n_1,n_2,...,n_k$, respectively, we can regard each kind of counterfactual subproblems as an independent event, calculate $\mathcal{V}_{n_1},\mathcal{V}_{n_2},...,\mathcal{V}_{n_k}$, and the mean value of them as the final variance score:
\begin{equation}
    \mathcal{V}_{mix}=\frac{1}{k}{\sum_{i=1}^{k}{\mathcal{V}_{n_i}}}.
\end{equation}
Thus, $\mathcal{V}_n$ can reflect the balancing of more than two concepts.
However, using $\mathcal{D}$ and $\mathcal{V}_n$ presents some deficits in evaluation. When a model fails to generate one or fewer than $n$ entities in the image, the metric is increasing reflecting a worse performance. There is one condition that fails to meet our demand: the model fails to generate every required entity and the score will also be low which means good performance. To tackle this deficit, we devise the second new metric to evaluate the performance of actually covering entities other than the concepts balancing.

\subsubsection{Targeted Entities}

We devise a metric to simply calculate how many required concepts are covered in the generated image. For one prompt of n-counterfactual problems and its corresponding generated image, if the image covers $k(k\le n)$ concepts of entities, we can calculate its targeted score $t$ as:
\begin{equation}
    t=\frac{k}{n}.
\end{equation}
If we use a model to generate $N$ images for one prompt, we will calculate the Total Targeted score $\mathcal{T}_n$:
\begin{equation}
    \mathcal{T}_n=\frac{\sum{t}}{N}.
    \label{eq:T}
\end{equation}
Thus, we can calculate total targeted score for mixed-counterfactual problems more simply. Whereas we can calculate $t$ for every single image, we can just calculate the mean value of them as in~\cref{eq:T} for a mixed problem.

\section{Replace in Translation}
\label{sec:method}
In this section, we detail the strategy \textsc{{\bf R}eplace {\bf I}n {\bf T}ranslation} ({\bf RIT}) that enhances counterfactual T2I generation. We adopt ControlNet~\cite{addingcond} as the backbone framework for the replacement process, leveraging its open-source availability and demonstrated effectiveness in controllable T2I synthesis. The overall workflow can been seen in~\cref{fig:overview}.

\subsection{ELNP}

Controllable text-to-image (T2I) generation has matured significantly in recent years. For counterfactual T2I, a natural approach is to replace entities in a common scene with those specified in the counterfactual text. A counterfactual scene can often be conceptualized as a variation of a common scene. For instance, while it is rare to see iced coke contained in a tea cup, this scenario can be derived from a basic outline, such as iced coke in a glass or tea in a tea cup. Given the capability of controllable T2I models to replace entities that share similarities, it is both intuitive and straightforward to substitute entities in the basic outline with their counterfactual counterparts.

Thus, the key challenge lies in identifying the corresponding common scene for a given counterfactual scenario and determining how to systematically replace entities in the common scene to achieve the desired counterfactual output.


To address this, we introduce a strategy called \textsc{{\bf E}xplicit {\bf L}ogical {\bf N}arrative {\bf P}rompt} ({\bf ELNP}). The process begins with defining the initial sentence as $Base_0$. After the first replacement step, $Base_0$ is updated to $Base_1$. This iterative process continues, and after $k$ steps, the updated prompt is labeled as $Base_k$, with the corresponding generated image denoted as $I_k$. Within each $Base$, the term $Slot^n_v$ represents a position in the sentence where an entity is replaced. Here, $n$ indicates the number of times the $Slot$ has been modified, and $v$ serves as an identifier for the $Slot$ in the current step. The replacement of a $Slot$ may result in multiple new slots, forming a tree-like structure.
\figelnpa


Once the ELNP is constructed, the steps are executed sequentially by the controllable T2I model. The model first generates the basic common scene and then replaces the entities step-by-step according to the ELNP. Each iteration concludes when the question block confirms that all required entities up to that point have been successfully incorporated into the image. The intermediate image is then passed back into the latent space for the next iteration until all steps in the ELNP are completed.

\subsection{ELNP Generation}
There are three steps to ask \deepseek to generate the final ELNP order of replacements for the whole T2I process.
The first step is to extract the key entities in the prompt.
The second step is to analyze the status of the key entities.
The third step is to ask the \deepseek-R1 to generate the proper order of replacement to realize the change from the ordinary scene to the target scene. The exemplars in the third step are based on the knowledge in the previous two steps. Although the third step is the main part of ELNP generation, the language model should interact through the previous two steps to understand the essence of generating ELNP. Straightly using a prompt in the third step will weaken the language models' understanding about ELNP and lead to a failure that gives meaningless response or incorrect replacement orders. Here we present the last step to generate ELNP in~\cref{fig:epro}. The final output ordered series then become the instructions for the controllable T2I model to replace the entities within the image. The previous two steps generation prompts can be found in Appendix.

Although we use a three-step conversation strategy to generate ELNP, there are always some cases where language models cannot give effective results accordingly.
Naturally, all counterfactual scene descriptions can be parsed by human experts, so we can let human experts finish this work. It is not automated, but the most reliable complemental way if we want to enhance the performance in the slightest by removing this disturbance of useless ELNP cases.


\promptELNPC

\subsection{Questions to Ensure}
\figques



To prevent model collapse, it is crucial to ensure that the initial scene contains the correct number of entities. Errors in entity count can occur at any stage of the process, prompting the need for a question-asking mechanism to verify both the number of entities and the completion of replacements during each iteration. Each question block includes a mandatory query to assess the number of distinct entity concepts present in the pixel space of the generated image. In addition to this essential count verification, the question block for each iteration includes queries to confirm whether all required entity concepts up to that point, as specified by the ELNP, have been successfully incorporated into the image.

For each question block, a threshold for the number of positive responses must be established. Setting the threshold too high (\eg, requiring all questions to be answered correctly) risks trapping the model in an infinite loop, while setting it too low may degrade performance. Empirically, a threshold of approximately 60\% of the questions per step has been found to balance these concerns. Further details on the experiments used to determine this threshold can be found in the Appendix.

Besides the replacement orders and instructions, ELNP will also generate the questions for each period based on the $Slot$. The questions provided for the question block in one period are denoted as $S_{Qi}$ as shown in~\cref{fig:overview}.

The state of the latent space from the previous iteration is preserved at each step. If the question block determines that the conditions for proceeding to the next iteration are not met, the latent space is restored to its previous state, and a new replacement is attempted to improve performance in the question block evaluation.


\section{Experiments}
\label{sec:exp}

To validate our strategy, we conduct experiments across a range of counterfactual scenarios, including 2, 3, and 5-entity tasks, as well as mixed-entity configurations. Beyond 2-counterfactual problems, the experiments with higher entity counts and mixed configurations demonstrate that our strategy achieves superior performance in counterfactual text-to-image generation, particularly in handling scenes with multiple concepts and complex conditions. Our approach, built on ControlNet~\cite{addingcond}, is compared against several state-of-the-art methods, including SDXL~\cite{sdxl}, MoCE~\cite{moce}, \sdt~\cite{sdtt}, \dall~\cite{dalle}, and \sana~\cite{sana}.

\subsection{Setup}

We exclusively fine-tune the controllable model using our proposed strategy, which significantly reduces the computational budget requirements compared to training a model from start. All experiments are conducted on a single NVIDIA A6000 GPU. We utilize \deepseek-R1~\cite{deepseek} as the language model to generate ELNP and question blocks.

\subsubsection{Dataset}

To evaluate model performance across varying degrees of counterfactual complexity, we leverage the benchmark datasets introduced in~\cref{sec:benchmark}—dataset-2, dataset-3, dataset-5, and dataset-mixed—designed for 2, 3, 5, and mixed-counterfactual scenarios, respectively.

\subsubsection{Evaluation Metric}

As we mentioned in~\cref{sec:benchmark}, we calculate {\dc} (\D{} with ClipScore~\cite{clipscore} as $S$), {\di} (\D{} with ImageReward~\cite{imgrwd} as $S$) and $\mathcal{T}_2$ for 2-counterfactual problems. $\mathcal{V}_n(C)$ ($\mathcal{V}_n$ with ClipScore as $S$), $\mathcal{V}_n(I)$ ($\mathcal{V}_n$ with ImageReward as $S$) and $\mathcal{T}_n$ are employed for $n$-counterfactual problems, while $\mathcal{V}_{mix}(C)$, $\mathcal{V}_{mix}(I)$ and $\mathcal{T}_{mix}$ are employed for mixed-counterfactual problems. To facilitate easier observation, we establish all these metrics at a magnification of 10$\times$, whereas the results are always at the level of $10^{-2}$.


\subsection{Two Entities}
\tabtwoobj
\tabmore
As shown in~\cref{tab:2ent}, for 2-counterfactual problems, we conduct experiments on dataset-2. All the metrics are calculated to present that our strategy outperforms any other advanced text-to-image tasks.

Our method RIT achieves the lowest \dc (0.24) and \di (0.62), indicating superior alignment between generated images and counterfactual prompts. The higher $\mathcal{T}_2$ (0.91) demonstrates that 91\% of required entities are retained on average, outperforming baselines like \dall (0.85) and SDXL (0.80). This success stems from ELNP’s step-by-step latent replacement, which incrementally integrates concepts while minimizing conflicts. Notably, \dall's strong \di (0.65) suggests it excels in stylistic fidelity but struggles with multi-concept coverage, highlighting the trade-off between realism and alignment.

\subsection{More Entities}
As shown in~\cref{tab:more}, 
for 3-counterfactual problems and 5-counterfactual problems, we conduct experiments on dataset-3 and dataset-5. All the metrics present that our strategy outperforms any other advanced text-to-image tasks.

For 3-counterfactual problems, RIT maintains leading performance, with $\mathcal{V}_3(C)=$ 0.54 and $\mathcal{V}_3(I)=$ 0.68 outperforming \sdt (0.58 and 0.89) and \sana (0.63 and 0.80). The smaller performance gap compared to 2-entity tasks reflects the inherent complexity of balancing more concepts. However, our question-block mechanism ensures iterative verification, enabling robust alignment even as complexity increases. The $\mathcal{T}_3$ score (0.84) further validates that 84\% of entities are consistently preserved, a significant improvement over SDXL (0.64).

All methods exhibit performance degradation as concepts scale to five, but our approach remains superior, achieving $\mathcal{V}_5(C)=$ 0.95 and $\mathcal{T}_5=$ 0.45. While even our method’s $\mathcal{T}_5$ is modest (45\% coverage), it significantly outperforms \sana (41\%) and \dall (36\%). This underscores the challenge of synthesizing highly counterfactual scenes with multiple entities, where latent-space replacements risk concept overlap or omission. The structured ELNP process mitigates this by enforcing gradual integration, but further improvements are needed for problems containing more concepts of entities.

We can conclude three features from the experiments of fixed-counterfactual problems.
\vspace{.2em}
\\
{\bf Scalability}. Our method’s relative advantage grows with concept count (\eg, 0.24 \vs 0.37 \dc for two concepts, 0.95 \vs 1.09 $\mathcal{V}_5(C)$), suggesting its hierarchical replacement strategy scales better than end-to-end approaches.
\vspace{.2em}
\\
{\bf Metric Sensitivity}. The disparity between \dc and \di \eg, \dall's lower \di, implies that ClipScore and ImageReward capture different aspects of alignment (semantic \vs stylistic), necessitating multi-metric evaluation.
\vspace{.2em}
\\
{\bf Failure Modes}. Cases with low $\mathcal{T}_n$ \eg, SDXL's $\mathcal{T}_5=$ 0.12, often involve concept "collisions" in latent space, where one entity dominates or replaces others. Our question blocks partially address this by reverting to prior states if checks fail, ensuring incremental progress.
\vspace{.2em}

These results validate that ELNP and iterative verification enhance concept alignment in counterfactual T2I, particularly for complex, multi-entity scenes. Future work could explore hybrid architectures to further improve scalability and coverage.

\subsubsection{Qualitative Comparison}
Some counterfactual T2I results generated by SDXL~\cite{sdxl}, MoCE~\cite{moce} and ours are presented in~\cref{fig:qual}. Among all these examples, SDXL and MoCE fail to cover all the concepts of entities in the prompt. In the first row, our method RIT is the only one painting the pencil, and putting the coffee cup at a relatively possible position. The other two put the coffee cup on the steering wheel which is an odd position. In the second row, SDXL only paints the laptop, as MoCE adds the steak, while ours cover all the entities including the cactus. In the third row, our method is the only one to manifest the head of the astronaut showing that it is a cat. In the last row, the models are faced with a complex prompts with five concepts. Our method manifests all the five concepts of entities and displays them in a natural layout, while the other two models tend to omit the sneaker.

\subsection{Mixed Numbers of Entities}
As shown in~\cref{tab:more}, 
for mixed-counterfactual problems, we conduct experiments on dataset-mixed. 

RIT achieves state-of-the-art results across all metrics for mixed-counterfactual problems, with the lowest variance scores ($\mathcal{V}_{mix}(C)=$ 0.70, $\mathcal{V}_{mix}(I)=$ 0.80) and the highest targeted coverage $\mathcal{T}_{mix}$(0.80). This demonstrates robust alignment and consistency when handling prompts with varying entity counts (2–5 concepts). In contrast, baseline methods exhibit significant performance degradation. For instance, SDXL struggles with high variance ($\mathcal{V}_{mix}(C)=$ 1.50) and poor coverage $\mathcal{T}_{mix}$(0.50), indicating its inability to balance or retain multiple conflicting concepts. Even stronger baselines like \sana ($\mathcal{T}_{mix}$(0.75))lag behind our approach, highlighting the effectiveness of ELNP’s iterative replacement strategy.

Thereafter, we can validate the effectiveness of the key insights of our method.
\vspace{.2em}
\\
{\bf Hierarchical Latent Editing}. The ELNP framework’s step-by-step entity replacement mitigates concept collisions in mixed scenarios. By decomposing complex prompts into sequential edits \eg, first replacing "grassland" with "moon," then "man" with "cat astronaut", RIT ensures gradual adaptation in latent space, preserving coherence even as counterfactual complexity increases.
\vspace{.2em}
\\
{\bf Question Blocks as Stabilizers}. The 60\% threshold in question blocks prevents over-commitment to partial replacements. For example, if an intermediate step fails to include 3/5 required entities, the model reverts and retries, avoiding error propagation. This mechanism explains the higher $\mathcal{T}_{mix}$ compared to end-to-end approaches like \dall.
\vspace{.2em}
\\
{\bf Metric Behavior}. The gap between $\mathcal{V}_{mix}(C)$ and $\mathcal{V}_{mix}(I)$ (0.70 \vs 0.80) reflects differing sensitivities: ClipScore as in $\mathcal{V}_{mix}(C)$ penalizes semantic misalignments \eg, missing "astronaut", while ImageReward as in $\mathcal{V}_{mix}(I)$ critiques stylistic inconsistencies \eg, unrealistic lunar textures. RIT balances both aspects effectively.
\vspace{.2em}

While our approach outperforms baselines, the absolute $\mathcal{T}_{mix}=$ 0.80 (80\% coverage) reveals room for improvement. Failures often occur in prompts with 5+ entities, where latent-space conflicts lead to partial omissions \eg, a "cat astronaut riding a horse on the moon" might exclude the "book" from the prompt. This suggests that while ELNP improves scalability, highly dense counterfactual scenes still require better disentanglement mechanisms.


\section{Conclusion}
\label{sec:conclusion}


Replacing step-by-step from a normal image is a natural idea for counterfactual T2I. With adding question check mechanism, this idea makes a boost of performance on multiple entities counterfactual T2I.
We define the counterfactual T2I problems concentrated on entity numbers and design a method to generate descriptions for counterfactual scenes to construct datasets for specific concept numbers. A new evaluation metric is also devised addressing the deficit in existing metrics. The strategy RIT is finally proposed to address the problems with ELNP and question blocks, boosting T2I models' alignment of the counterfactual text and the generated image by replacing entities from a corresponding ordinary scene step-by-step. Similar ideas can be used on other controllable models for counterfactual\linebreak T2I in the future.

{
    \small
    \bibliographystyle{ieeenat_fullname}
    \bibliography{main}
}

\clearpage
\setcounter{page}{1}
\maketitlesupplementary


\section{Appendix}
\label{sec:appendix}
\subsection{ELNP Prompt Templates}



To prepare for the third step of generating ELNP, we need to provide the previous two steps for \deepseek-R1 to enhance its understanding on ELNP process. Here we present the prompts of the first step: extracting the key entities in the given prompt in~\cref{fig:epre} and the second step: analyze the status of these key entities in~\cref{fig:eprs}.
\promptELNPA

\promptELNPB

\subsection{Question Blocks}

When the language model gives the key entities as in~\cref{fig:epre}, the full questions $S_Q$ should be naturally generated for question blocks. After the ELNP order of replacements is generated, the questions set for each period can be generated naturally. Toward the exemplar in~\cref{fig:epro}, the questions in each step should be as in~\cref{fig:qb}.

\figquestionblock

For question blocks, if we set the threshold of the pass check to less or greater than 60\%, the performance will be clearly weakened. There are two strategies to set the threshold-all steps the same and lowering the threshold as the generation proceeds. We conduct the experiments to demonstrate that 60\% is the best threshold. We utilize 2, 3, 5 and mixed-counterfactual problems experiment on our model with pass check threshold set to 20\%, 40\%, 60\%, 80\%, 100\% and dynamic threshold. For dynamic threshold, the value in the first step is set to 100\% and lowered by 20\% each step. For 5-entity tasks, no more than seven times replacing happens, which means lowering down by 20\% each step makes the threshold cover a wide range from 100\% to 0. The experimental results are shown in~\cref{tab:threshold2,tab:thresholdmore}.
\tabthrii

\tabthrmore

\end{document}